\pdfoutput=1
\documentclass[conference]{IEEEtran}
\usepackage{cite}
\usepackage{amsmath,amssymb,amsfonts}
\usepackage{graphicx}
\usepackage{textcomp}

\usepackage{times}
\usepackage{epsfig}
\usepackage{bm}
\usepackage{booktabs}

\usepackage{amsfonts}
\usepackage{algorithm}
\usepackage[noend]{algpseudocode}
\algnewcommand{\LineComment}[1]{\State \(\triangleright\) #1}

\usepackage{subcaption}
\usepackage{hyperref}

\def\etal{\emph{et al}.}

\def\BibTeX{{\rm B\kern-.05em{\sc i\kern-.025em b}\kern-.08em
    T\kern-.1667em\lower.7ex\hbox{E}\kern-.125emX}}

\IEEEoverridecommandlockouts
    
\begin{document}

\title{Multimorbidity Content-Based Medical Image Retrieval Using Proxies
\thanks{Corresponding author: Yunyan Xing (e-mail: yunyan.xing@monash.edu).}}

\author{
    \IEEEauthorblockN{Yunyan Xing$^1$, Benjamin J. Meyer$^1$, Mehrtash Harandi$^1$, Tom Drummond$^2$, and 
    Zongyuan Ge$^{3,4}$}\\
    \IEEEauthorblockA{$^1$Department of Electrical and Computer Systems Engineering, Monash University, Australia}
    \IEEEauthorblockA{$^2$Melbourne Connect, University of Melbourne, Australia}
    \IEEEauthorblockA{$^3$Monash-Airdoc Research, Monash University, Australia}
    \IEEEauthorblockA{$^4$Monash Medical AI, Monash eResearch Centre, Monash University, Australia}
}

\maketitle

\begin{abstract}
Content-based medical image retrieval is an important diagnostic tool that improves the explainability of computer-aided diagnosis systems and provides decision making support to healthcare professionals.
Medical imaging data, such as radiology images, are often multimorbidity; a single sample may have more than one pathology present. As such, image retrieval systems for the medical domain must be designed for the multi-label scenario.
In this paper, we propose a novel multi-label metric learning method that can be used for both classification and content-based image retrieval. In this way, our model is able to support diagnosis by predicting the presence of diseases and provide evidence for these predictions by returning samples with similar pathological content to the user. In practice, the retrieved images may also be accompanied by pathology reports, further assisting in the diagnostic process.
Our method leverages proxy feature vectors, enabling the efficient learning of a robust feature space in which the distance between feature vectors can be used as a measure of the similarity of those samples. Unlike existing proxy-based methods, training samples are able to assign to multiple proxies that span multiple class labels. This multi-label proxy assignment results in a feature space that encodes the complex relationships between diseases present in medical imaging data.
Our method outperforms state-of-the-art image retrieval systems and a set of baseline approaches. We demonstrate the efficacy of our approach to both classification and content-based image retrieval on two multimorbidity radiology datasets.
\end{abstract}

\section{Introduction}

Radiology is a vital tool for the diagnosis of disease. With the demand for medical imaging increasing rapidly \cite{hosny2018artificial}, computer-aided diagnosis systems can help to improve the radiology workflow.
Two useful computer-aided diagnosis tasks are pathology classification and Content-Based Image Retrieval (CBIR), i.e. the process of searching an image database for samples that are pathologically similar to a query image.
In practice, such a medical CBIR system may also return pathology reports alongside the retrieved samples. 
Returning similar images and their pathology information to the user provides evidence and context for the pathology classifications made by the system. This can in turn help foster trust between healthcare professionals and the computer-aided diagnosis tool.
Further, a classification and retrieval system can help reduce the workload of healthcare professionals by assisting in the generation of radiology reports \cite{haq2021deep} and help to reduce the high inter-observer variability that occurs when analysing radiology images \cite{chen2018order}.

\begin{figure}[t!] 
    \centering
    \includegraphics[width=0.47\textwidth]{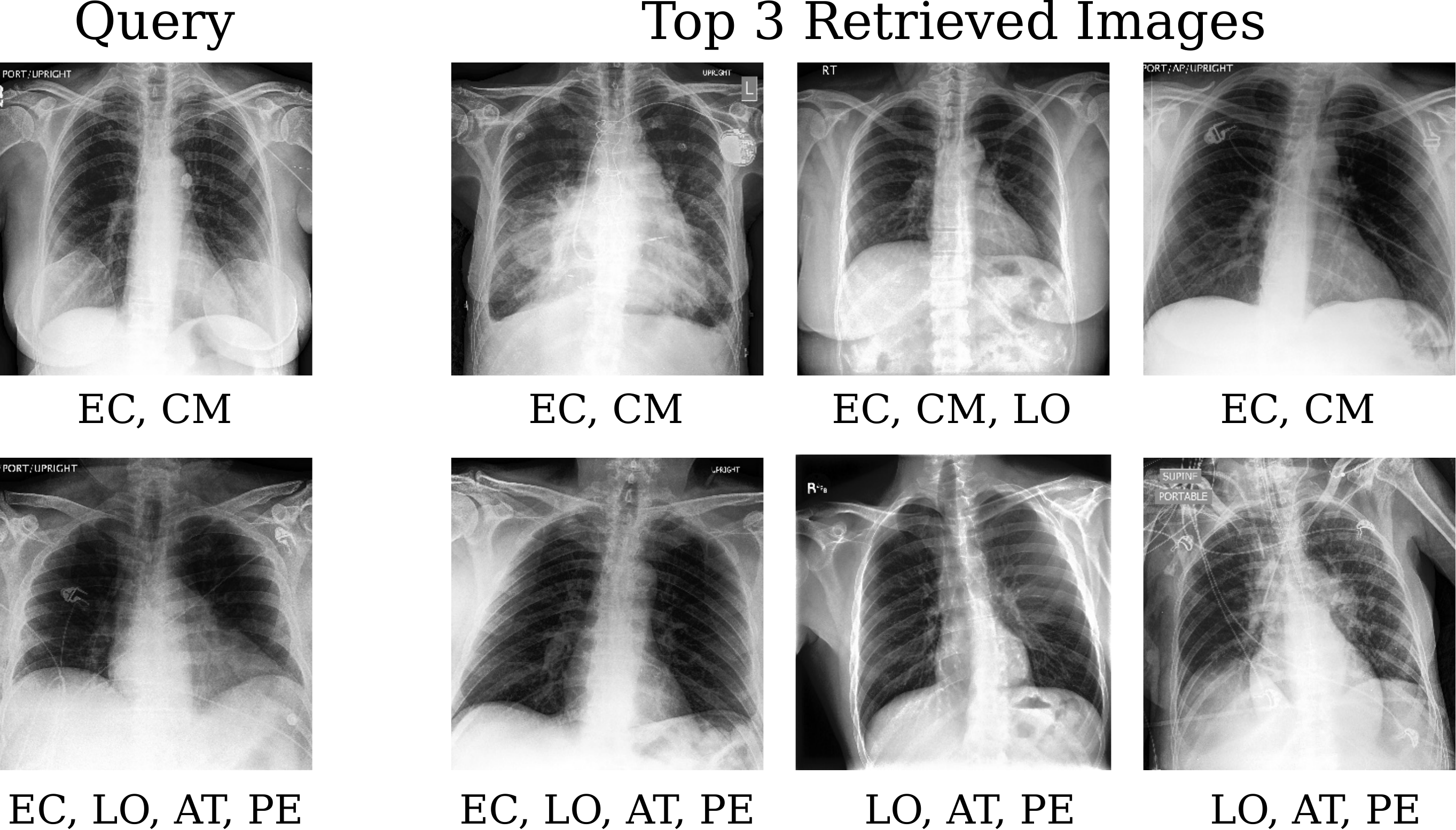}
    \caption{Example X-ray retrieval results. Images are annotated with disease labels; see Section \ref{sec:datasets} for label definitions.}
    \label{fig:key_result}
\end{figure}

Deep metric learning methods learn a feature space in which distance is a measure of similarity. As such, metric learning is well suited to the problem of content-based image retrieval. The feature spaces learned by metric learning approaches have been shown to generalise well to unseen samples \cite{oh2016deep, sohn2016improved, harwood2017smart, meyer2018deep}, with demonstrated efficacy to retrieval \cite{cakir2019deep, zhong2021deep}.
Conventional deep classification models often require vast quantities of high-quality annotated data in order to be accurately trained \cite{marcus2018deep}. Metric learning models can reduce over-fitting \cite{li2021deep}, resulting in better performance in few shot learning \cite{sung2018learning, li2020revisiting} and data limited scenarios \cite{meyer2018deep}. This is particularly important in the medical domain, as some diseases may be rare and annotating data has a high associated cost.

One family of metric learning approaches are proxy-based methods \cite{movshovitz2017no, teh2020proxynca++}. Proxies are trainable model parameters that are used to approximate the real distribution of training set feature vectors.
Using proxies enables efficient model training, as training samples need only be compared to the relatively small number of proxies, rather than to one another.
Proxy methods are able to learn faster than other metric learning methods and result in a more generic feature space than commonly used triplet-based methods \cite{movshovitz2017no}.
However, existing proxy methods are not designed for multi-label data and cannot be directly applied to multimorbidity CBIR.

In this paper, we propose a novel proxy-based metric learning method for multimorbidity computer-aided diagnosis. Existing proxy approaches generally allow a sample to assign to just a single proxy during training. Unlike these approaches, our method is explicitly designed for multi-label data by allowing X-rays with multiple disease findings to assign to all relevant proxies. This results in a metric feature space that encodes the complex interactions between different diseases. 
Unlike existing methods that tend to define a single proxy per class label, we propose the use of several proxies per disease class, allowing the semantic variations that exist within the distribution of a single disease to be encoded in the feature space. In this way, our method supports both intra-class and inter-class multi-proxy assignment.
Further, we propose the use of \textit{negative proxies}. By treating negative samples (i.e. those with no positive labels) as their own training class with proxies, a feature space is learned that better encodes the relationships between X-ray samples with no disease findings.

While many metric learning approaches do not perform well for classification \cite{rippel2015metric}, our method optimises a classification loss directly on the feature vectors extracted by a deep neural network. This results in a trained model that can be successfully applied to both pathology classification and content-based image retrieval. 
The major contributions of this paper can be summarised as follows:
\begin{itemize}
    \item We propose a novel proxy-based metric learning algorithm for multi-label data (Section \ref{sec:proxybce}).
    \item We introduce negative class proxies for encoding the important relationships between X-ray samples with no disease findings (Section \ref{sec:neg_proxy}).
    \item We propose defining multiple proxies for each class label and demonstrate the performance benefit (Sections \ref{sec:proxybce} and \ref{sec:num_proxies}).
    \item We demonstrate that our method outperforms conventional deep classification models (Section \ref{sec:chexpert_res}).
    \item We show that our approach achieves state-of-the-art CBIR performance on multimorbidity radiology datasets (Sections \ref{sec:nih_res} and \ref{sec:chexpert_res}).
\end{itemize}

\section{Related work}
\subsection{Content-based medical image retrieval} \label{related:cbir}
As medical imaging data is often multimorbidity, i.e. a single sample may show the presence of several pathologies, we focus on the problem of multi-label CBIR. Medical image retrieval remains a challenging problem, owing to the often minuscule visual differences between pathologies, as well as the existence of complex relationships between pathologies \cite{zhang2016large, li2018large, haq2021deep}. 
Existing medical image retrieval systems include those based on handcrafted features and shallow methods \cite{rahman2011learning, wang2012semi, gong2012iterative, lan2018simple}, as well as deep learning methods \cite{shah2016deeply, liu2016generating, anavi2015comparative, chen2018order, conjeti2017hashing, liu2016deep, erin2015deep, haq2021deep}.

In the deep learning domain, multi-label metric learning and hashing is a common approach \cite{chen2018order,conjeti2017hashing,liu2016deep,erin2015deep}. Chen \etal \cite{chen2018order} propose a method that optimises a combination of ranking loss and multi-label classification loss, while Conjeti \etal \cite{conjeti2017hashing} introduce a Deep Residual Hashing network that incorporates a retrieval loss and regularisation techniques to improve CBIR performance. A pair-wise Deep Supervised Hashing method is proposed by Liu \etal \cite{liu2016deep}, whereby images are mapped to discreet codes, allowing Hamming distance to be used as a measure of the similarity between samples. Further discussion on multi-label metric learning is found in Section \ref{related:mlml}.

Taking a different deep learning approach, Haq \etal \cite{haq2021deep} leverage a conventional Convolutional Neural Network (CNN) multi-label classifier trained with binary cross-entropy. A community-based graph structure is proposed for efficient search in large retrieval databases. As demonstrated in our experiments (Section \ref{sec:exp}), a conventional CNN classifier is limited in its ability to encode rich semantic relationships in the feature vector space. This results in poorer CBIR performance compared to our multi-label metric learning approach that directly optimises the feature space. 

\subsection{Metric learning}
The aim of metric learning is to learn a feature space in which standard distance measures, such as Euclidean distance, can be used as a measure of similarity. For example, one would expect the feature vectors belonging to sample images with similar semantic content to be located nearby, while those from images with dissimilar semantic content to be located further apart. Metric feature spaces have applications including retrieval \cite{gao2014soml}, ranking \cite{cakir2019deep}, out-of-distribution detection \cite{meyer2019importance} and novel class image generation \cite{ditria2020opengan}. 

\subsubsection{Triplet Methods} \label{related:triplet}

One of the main approaches to metric learning is derived from Siamese networks \cite{bromley1993signature} with contrastive loss \cite{hadsell2006dimensionality, chopra2005learning}, whereby positive pairs of images (with matching semantic content) and negative pairs (with non-matching semantic content) are passed through the same network. The network parameters are updated such that the feature vectors of positive pairs are pulled together, while those of negative pairs are pushed apart. An improvement on pairwise metric learning methods are triplet-based methods \cite{weinberger2005distance}, which construct trios of training images containing two samples with matching semantic content and one sample with differing semantic content. Triplet loss attempts to pull the positive pair of samples closer together than the anchor positive sample and the negative sample, by a set margin.

Metric learning literature often aims to improve triplet methods by performing intelligent mining of ``hard" triplets \cite{schroff2015facenet, harwood2017smart}. While other streams of literature aim to generalise triplet loss, such as by allowing multiple comparisons within a single mini-batch \cite{sohn2016improved} or by employing a lifted structured embedding \cite{oh2016deep} that allows computation between every positive and negative pair in the batch. Local triplet loss can also be combined with a global loss term to improve performance \cite{kumar2016learning}.
Triplet-based metric learning has shown good performance in embedding learning problems and extreme classification, but is often poor at regular classification problems compared to conventional neural network classifiers \cite{rippel2015metric}. Further, triplet methods suffer from computational bottlenecks in terms of triplet mining.

\subsubsection{Neighbourhood and Proxy Methods}

Analysing a larger neighbourhood of samples at each training iteration can allow for more efficient updates to the model and a more robust distance metric \cite{meyer2018deep}. Neighbourhood Component Analysis (NCA) \cite{goldberger2004neighbourhood} considers all nearby samples, minimising a probabilistic loss based on a sum of Gaussian distances within the neighbourhood. As the feature vectors of every sample change after each training iteration, it is computationally infeasible to minimise this loss exactly with a deep neural network. To make training practical, a cache of training feature vectors can be stored and periodically updated, allowing an approximation of NCA loss to be optimised \cite{meyer2018deep}. 

Another approach to making neighbourhood methods computationally feasible is to employ proxy feature vectors \cite{movshovitz2017no}. Proxy features are trainable model parameters that are assigned a class label and used as a proxy for real training feature vectors belonging to that same class. By minimising a proxy-based version of NCA (Proxy-NCA \cite{movshovitz2017no, teh2020proxynca++}), training features need only be compared to the proxy features, rather than to each other. As the number of proxies is generally set to equal the number of class labels, this significantly reduces the computational complexity. Kim \etal \cite{kim2020proxy} propose a loss function with advantages of both pair-wise methods and proxy-based methods by assigning samples to proxies using sample-to-sample relations. Similarly, N-pair loss is implemented using proxies by Aziere \etal \cite{aziere2019ensemble}, while SoftTriple loss \cite{qian2019softtriple} improves the representation of intra-class variations using a method analogous to proxies.
In this work, we propose a novel multi-label proxy metric learning method that allows training samples to assign to multiple inter-class and intra-class proxies. This results in a model that can be applied to both multi-label classification and multi-label image retrieval.

\subsubsection{Multi-label Metric Learning} \label{related:mlml}

Beyond the discussed medical CBIR approaches \cite{chen2018order,conjeti2017hashing,liu2016deep,erin2015deep} (Section \ref{related:cbir}), other multi-label metric learning approaches include the multi-label extension of triplet loss by Sumbul \etal \cite{sumbul2021informative}. A two-step triplet sampling algorithm is proposed, that uses multi-label similarity to select a diverse set of triplets for a training mini-batch.
Annarumma \etal \cite{annarumma2018deep} propose a multi-label triplet method that selects several positive examples for each training sample, such that each of the sample's positive class labels are present in the constructed positive pairs. 
As multi-label extensions of triplet learning, these methods inherit the inefficiencies of triplet metric learning discussed in Section \ref{related:triplet}, in terms of high computational complexity, limited ability to encode complex intra-class and inter-class relationships, and poor classification performance. 
To avoid these known problems with triplet loss, our work focuses on computationally efficient proxies and directly minimises a distance-metric based classification loss that is effective for both classification and content-based image retrieval.
Beyond multi-label triplet methods, Li \etal \cite{li2019reconstruction} optimise a two-way distance metric loss between image and label embeddings, both extracted by neural networks. This method uses the entire neighbourhood of feature vectors to compute the loss. Avoiding such inefficient neighbourhood analysis is a primary motivator behind our use of proxies.

Further to not being proxy methods, the discussed existing approaches do not give special consideration to negative samples (i.e. examples with no positive labels). Such samples are extremely common in medical data (e.g. a radiology image with no disease finding). Recognising this, our method is explicitly designed for such data via the introduction of negative proxies (Section \ref{sec:neg_proxy}).
Additionally, our approach was designed for the consolidated dual use case of medical image retrieval and pathology classification, while most existing multi-label metric learning methods were designed only for a single use case.
We quantitatively evaluate our approach against existing multi-label metric learning methods \cite{chen2018order,liu2016deep,erin2015deep} in Section \ref{sec:nih_res}.

\begin{figure*}[t!] 
    \centering
    \begin{subfigure}[t]{0.85\textwidth}
        \centering
        \includegraphics[width=\textwidth]{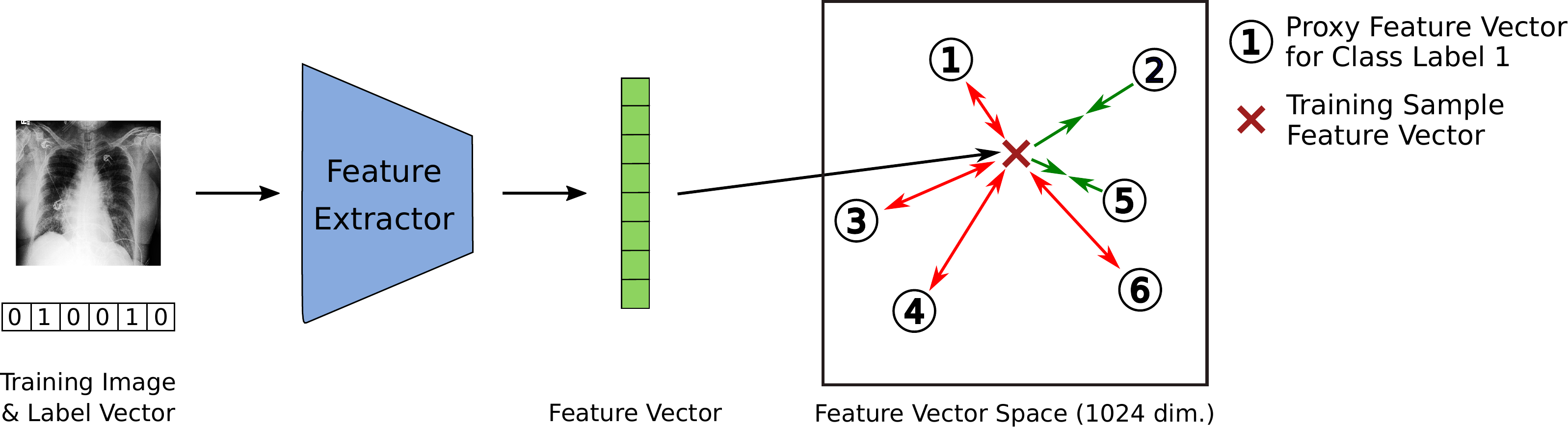}
        \caption{Training the system.}
    \end{subfigure}
    \par\bigskip
    \begin{subfigure}[t]{0.85\textwidth}
        \centering
        \includegraphics[width=\textwidth]{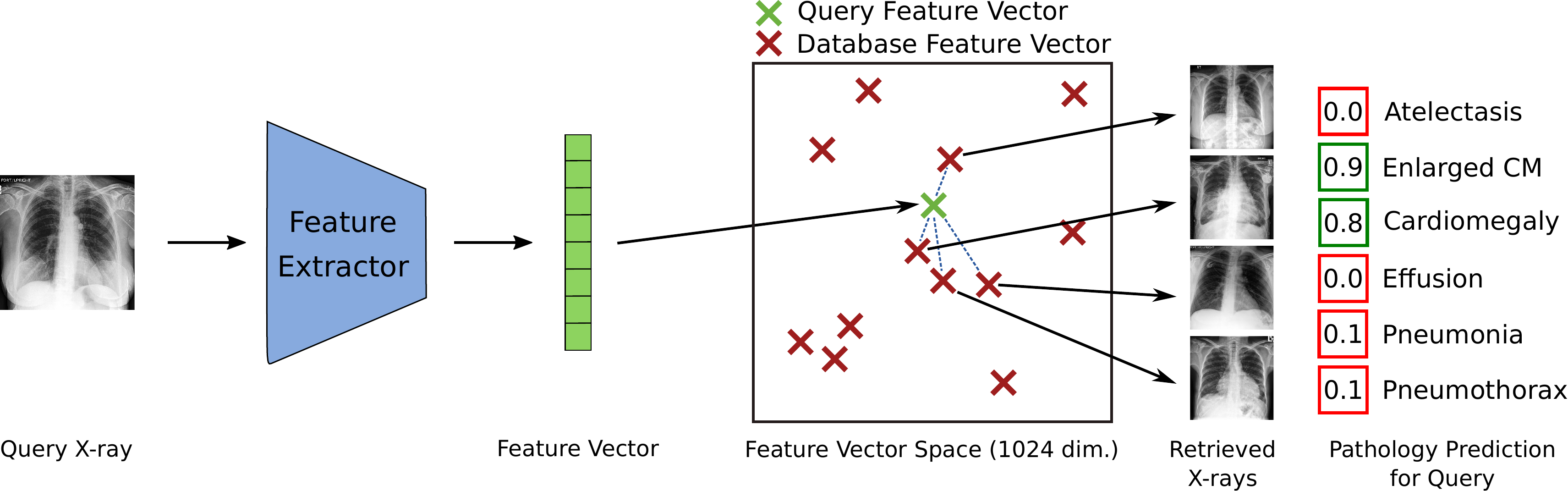}
        \caption{Querying the system.}
    \end{subfigure}
    \caption{Overview of training and testing of our system.}
    \label{fig:overview}
\end{figure*}

\subsubsection{Using Multiple Proxies per Class}
Further to enabling multi-label proxy metric learning, we propose to improve the representation of intra-disease variations by defining multiple proxies for each disease class. 
Although existing works have proposed to define multiple proxies for a single class \cite{qian2019softtriple,liu2021deep,rippel2015metric}, these methods differ significantly from our approach.
Qian \etal \cite{qian2019softtriple} propose the use of multiple centroids for each class, but a single weighted per-class centroid is used in the loss function, limiting the intra-class variations that can be captured. Conversely, our loss function incorporates all proxies (both intra-class and inter-class) independently.
Similar to approximate NCA methods \cite{meyer2018deep}, Liu \etal \cite{liu2021deep} couple a proxy with each data instance, which is by definition different to the dataset approximating proxies in our method. The proxies used in our proposed approach are trained model parameters that are efficient approximations of the data distribution.
Although not strictly proxies, Rippel \etal \cite{rippel2015metric} use $k$-means clustering and penalise overlapping clusters from different classes, with multiple $k$-means centres defined for each class. In this sense, the centres can be considered proxies of the data distribution. However, neither approaches in Rippel \etal \cite{rippel2015metric} and Liu \etal \cite{liu2021deep} have the efficiency benefits of the parameterised proxies used in our paper.

\section{Method}
An overview of our approach is shown in Fig. \ref{fig:overview}. Given a query image, such as a chest X-ray, our method is able to predict the presence of multiple diseases, as well as return a set of semantically similar X-rays from a retrieval database to the user. The similarity of samples is measured by Euclidean distance in a feature vector space. Features are extracted by a convolutional neural network that is trained with our novel multi-label metric learning method.

\subsection{Metric learning problem statement}
Let $\mathcal{X}=\{\mathbf{x}_1, \mathbf{x}_2, \mathbf{x}_3, ... \mathbf{x}_n\}$ be a set of $n$ images from a multi-label dataset containing $c$ class labels. The corresponding set of ground truth label vectors is $\mathcal{Y}=\{\mathbf{y}_1, \mathbf{y}_2, \mathbf{y}_3, ... \textbf{y}_n\}$ where $\mathbf{y}_i \in \{ 0, 1 \}^c$ is the label vector for the $i$-th image. A value of 1 at location k in $\mathbf{y}_i$ indicates the presence of the k-th class in image $\mathbf{x}_i$.
We aim to learn a distance metric $d$, such that: 
\begin{equation}
    d(\mathbf{x}_i, \mathbf{x}_j) \propto \|\mathbf{y}_i - \mathbf{y}_j\|_2.
\end{equation}

Such a distance metric is achieved by learning a transformation from the image space to a feature vector space in which the Euclidean distance between features can be used as a measure of similarity, i.e.:
\begin{equation}
    d(\mathbf{x}_i, \mathbf{x}_j) = \|f(\mathbf{x}_i) - f(\mathbf{x}_j)\|_2,
\end{equation}
where $f$ represents a neural network encoder and $f(\mathbf{x}_i)$ is the feature vector that is output from the network for image $\mathbf{x}_i$.
Given a well learned metric feature vector space, we would expect that the distance between similar samples in the feature space will be small, while the distance between dissimilar samples will be large.

\subsection{Proxy feature vectors}
The feature vector extracted from image $\mathbf{x}$ by the model $f$ is defined as:
\begin{equation}
    \mathbf{v} = f(\mathbf{x}),
\end{equation}
where $\mathbf{v} \in \mathbb{R}^{d}$ and $d$ is the dimensionality of the feature vector space, realised by the model $f$. In general, the feature space may have any dimensionality; in our experiments $f$ produces features with 1024 dimensions. A proxy $\mathbf{p} \in \mathbb{R}^{d}$ is defined as a trainable feature vector that is assigned a class label and represents the set of, or a subset of, the real training sample feature vectors that belong to that class. During training, sample feature vectors are compared with the proxies rather than with the much larger number of other training samples. In this way, these trainable parameters act as a \textit{proxy} for the real distribution of training samples.

\subsection{Multi-label metric learning with proxies} \label{sec:proxybce}
Existing proxy-based metric learning methods \cite{movshovitz2017no,teh2020proxynca++} are designed for datasets with a single positive class label per sample. Generally, these methods define a single proxy for each class label, and training samples are only able to be assigned to one proxy.
Since medical imaging data is often multimorbidity, we propose a novel proxy-based metric learning approach for multi-label datasets. In this approach, training samples are able to assign to multiple proxies spanning multiple class labels. We also generalise our method to allow for the definition of multiple proxies per class. Having more than one proxy for a single class may allow for intra-class variations to be better captured by the model, as well as more complex interactions between class labels. This means that a sample's multi-proxy assignment may occur both inter-class and intra-class.

We define a set of $m$ proxies for each $c$ class labels as:
\begin{equation*}
    \mathcal{P} = \begin{Bmatrix}
        \mathbf{p}_{1,1} & \mathbf{p}_{1,2} & \dots & \mathbf{p}_{1,c}\\
    \mathbf{p}_{2,1} & \mathbf{p}_{2,2} & \dots & \mathbf{p}_{2,c}\\
    \vdots & \vdots& \ddots & \vdots\\
    \mathbf{p}_{m,1} & \mathbf{p}_{m,2} & \dots & \mathbf{p}_{m,c}
    \end{Bmatrix}
\end{equation*}
where each $\mathbf{p}_{i,j} \in \mathbb{R}^{d}$ corresponds to the $i$-th proxy feature for the $j$-th class label.

During training, we optimise a multi-label proxy-based loss that allows training features to assign to multiple proxies. The multi-proxy assignment happens both intra-class (when $m > 1$) and inter-class, when a training sample has more than one positive label (i.e. when $\|\mathbf{y}\|_0 > 1$, where the 0-norm of $\mathbf{y}$ is the number of non-zero elements).

The loss function $\mathcal{L}$ for sample $\mathbf{x}$ with labels $\mathbf{y}$ is defined in (\ref{loss_function}). To deal with the large imbalance of positive and negative occurrences of classes that is common in medical imaging data \cite{irvin2019chexpert, wang2017chestx}, per-class positive and negative weights are used in the loss function. With $n_{j+}$ being the total number of positive class $j$ samples and $n_{j-}$ being the number of negative class $j$ samples, the positive and negative weights for class $j$ are defined as $w_{j+} = n_{j-}/(n_{j+} + n_{j-})$ and $w_{j-} = n_{j+}/(n_{j+} + n_{j-})$, respectively. Before calculating the loss, feature vectors and proxies are normalised. The hyperparameter $\sigma$ is a fixed value that sets the width of the Gaussian windows in our multi-label proxy loss function. This value is the same for all proxies.

\footnotesize
\begin{multline}\label{loss_function}
    \mathcal{L}(\mathbf{v},\mathbf{y}; \mathcal{P}) =  - \sum_{j=1}^c \left( w_{j+} y_j \log \left( \frac{1}{m}\sum_{i=1}^m\exp{\frac{-\|\mathbf{v}-\mathbf{p}_{i,j}\|^2}{2\sigma^2}} \right) \right. \\
     \left. - w_{j-}(1-y_j) \log \left( 1-\frac{1}{m}\sum_{i=1}^m\exp{\frac{-\|\mathbf{v}-\mathbf{p}_{i,j}\|^2}{2\sigma^2}} \right) \right) .
\end{multline}
\normalsize

\subsection{Negative proxies} \label{sec:neg_proxy}
Training samples that are negative for all labels, e.g. healthy chest X-rays with no disease findings, can be considered as their own class for model training purposes. Formally, an X-ray is considered a \textit{negative sample} when $\|\mathbf{y} \|_0 = 0$, where the 0-norm calculates the number of non-zero elements in $\mathbf{y}$. An additional element can be concatenated to the label vector to represent negative samples, i.e. $\mathbf{y} \oplus \delta(\| \mathbf{y} \|_0 = 0)$, where $\oplus$ denotes concatenation and $\delta(.) \in \{0, 1\}$ is an indicator function. As negative samples are now represented as a class label, proxies must also be defined for this new class. Training with additional proxies for negative samples can result in a better structured feature space, as negative samples will cluster together around those proxies. Without such proxies, the only training constraint for negative samples is for them to be located far away from positive class proxies. However, there is no loss term that encourages negative samples to be located nearby one another, despite their semantic similarity. The inclusion of proxies for negative samples, which we name \textit{negative proxies}, introduces a training constraint that negative samples should be located nearby in the feature space. As shown in our experiments in Section \ref{sec:neg_proxy_res}, negative proxies result in both better classification and CBIR performance.

% -------------------------------Algorithm--------------------------
% {\centering
% \hspace{0.5cm}
% \begin{figure}
% \centering
% \begin{minipage}[!t]{\linewidth}
\begin{algorithm}[t]
\caption{Training procedure.}\label{alg:cap}
% \begin{multicols}{2}
\begin{algorithmic}[1]
\Require \phantom{3} \newline Model $f$ with parameters $\bm{\theta}_f$;
\newline Proxies $\mathcal{P}$;
\newline Dataset $\mathcal{X,Y}$.

\While{not converged}
\State Sample $(\mathbf{x, y}) \sim \mathcal{X,Y}$ \Comment{Sample training data.}
\State $\mathbf{v} \gets f(\mathbf{x})$ 
\State $\mathbf{v} \gets \frac{\mathbf{V}}{\|\mathbf{V}\|_2}$ \Comment{Normalise feature vector.}
\For{each $\mathbf{p}_{i,j} \in \mathcal{P}$}
\State $\mathbf{p}_{i,j}\gets \frac{\mathbf{p}_{i,j}}{\|\mathbf{p}_{i,j}\|_2}$ \Comment{Normalise proxies.}
\EndFor
\Statex \Comment{Concatenate ($\oplus$) label to represent negative samples.}
\State $\tilde{\mathbf{y}} \gets \mathbf{y} \oplus \delta(\| \mathbf{y} \|_0 = 0) $
\State $\bm{\theta}_f \gets \bm{\theta}_f - $Adam$(\nabla \mathcal{L}(\mathbf{v}, \tilde{\mathbf{y}}; \mathcal{P})$
\State $\mathcal{P} \gets \mathcal{P} - $Adam$(\nabla \mathcal{L}(\mathbf{v}, \tilde{\mathbf{y}}; \mathcal{P})$
\EndWhile
\end{algorithmic}
% \end{multicols}
\end{algorithm}
% \end{minipage}
% \end{figure}
% \par
% }

\subsection{Training algorithm}
The training procedure for our approach is shown in Algorithm \ref{alg:cap}. Image and label pairs are sampled from the training dataset and feature vectors are extracted from the images. Both the feature vectors and the proxies are then constrained to a unit sphere using L2-normalisation. The label vector is modified to include an extra dimension that represents the negative (healthy) class. This extra label is set to a value of one when all other elements of the label vector are zero, otherwise the negative label is set to a value of zero. The extra label dimension is needed due to the inclusion of the negative proxies, as discussed in Section \ref{sec:neg_proxy}. The loss is then calculated according to (\ref{loss_function}), and model parameters and proxies are updated using the Adam optimisation algorithm \cite{kingma2014adam}.

\subsection{Multi-label classification}
We perform multi-label classification by analysing the similarity between a sample's feature vector and each of the proxy feature vectors. The classification score for disease label $j$ is:
\begin{equation}
    \mathrm{score}_j = \max_i\exp{\frac{-\|\mathbf{v}-\mathbf{p}_{i,j}\|^2}{2\sigma^2}}
\end{equation}
where $0\leq \mathrm{score}_j\leq1$. For each class $j$, we calculate the distance between the feature vector $\mathbf{v}$ and each of the proxies belonging to class $j$. The classification score is then calculated based on the distance to the nearest proxy of that class, where a score close to 1 indicates a high likelihood that disease $j$ is present in the sample image, while a score close to 0 indicates a low likelihood. A prediction $\mathrm{pred}_j$ for class $j$ can then be made by comparing the classification score to a discrimination threshold for that class $t_j$, as shown in (\ref{eq:pred}).

\begin{equation} \label{eq:pred}
    \mathrm{pred}_j = 
    \begin{cases}
      1 & \text{if $\mathrm{score}_j > t_j$}\\
      0 & \text{otherwise}
    \end{cases} 
\end{equation}

\subsection{Content-based image retrieval}
Given a well structured metric feature space, we expect the feature vectors of samples with similar pathology information to be located nearby. As such, we perform content-based image retrieval for a query image $\mathbf{q}$ by returning the $k$ database images corresponding to the $k$ feature vectors that are nearest to the query sample's feature vector. The image retrieval database, i.e. the set of images from which samples are retrieved based on a  query, is constructed from the labelled training samples.
The image retrieval procedure is outlined in Algorithm \ref{alg:cbir}. The feature vector distance between the query sample and each of the samples in the database are calculated. The database samples are then sorted based on the distance to the query sample, in ascending order. Finally, the $k$ images that are most similar to the query image are returned to the user as the output of the CBIR system.

% -------------------------------Algorithm--------------------------
% {\centering
% \hspace{0.5cm}
% \begin{figure}
% \centering
% \begin{minipage}[!t]{\linewidth}
\begin{algorithm}[t]
\caption{CBIR of $k$ images for query image $\mathbf{q}$.}
% \begin{multicols}{2}
\begin{algorithmic}[1]
\Require \phantom{3} \newline Trained model $f$;
\newline Retrieval database $\mathcal{X,Y}$.

\State Initialise: $\mathrm{dist} \gets$ \O, $R \gets$ \O \Comment{$R$ retrieved images.}
\For{each $\mathbf{x}_i \in \mathcal{X}$}
\State $d \gets \|f(\mathbf{q}) - f(\mathbf{x}_i)\|$ \Comment{Dist. b/w query and samples.}
\State $\mathrm{dist} \gets \mathrm{dist}\cup{d}$ \Comment{Add $d$ to set of distances.}
\EndFor
\State $\mathrm{idx} \gets $argsort$(\mathrm{dist}, $ ascent$)$ \Comment{Sort indices of $\mathrm{dist}$.}
\State $i \gets 1$
\While{$i \leq k$} \Comment{Pick k most similar images.}
\State $R \gets R\cup{\mathbf{x}_{\mathrm{idx}(i)}}$ \Comment{Add next similar image to $R$.}
\State $i \gets i+1$
\EndWhile
\State return $R$
\end{algorithmic}
% \end{multicols}
\label{alg:cbir}
\end{algorithm}
% \end{minipage}
% \end{figure}
% \par
% }

\subsection{Baselines for evaluation} \label{sec:baselines}
We compare our proposed approach to seven state-of-the-art CBIR methods from the literature \cite{wang2012semi, gong2012iterative, erin2015deep, liu2016deep, chen2018order, lan2018simple, haq2021deep}, including CNN classifier methods, feature-based deep learning methods and hashing methods. For further evaluation, we train appropriate baseline models to benchmark our approach against. These baselines are described below. For fairness, our method uses the same base network architecture as all baselines methods, as well as the highest performing evaluated method from literature \cite{haq2021deep}. 

\subsubsection{Multi-label classifier (\textit{DenseNet w/ BCE}).} This method is a conventional CNN classifier, trained with multi-label binary cross entropy loss. The head of the network is a fully connected layer that outputs class-wise prediction scores. To extract feature vectors for image retrieval, we bypass the final fully connected layer, resulting in a feature vector with the same dimension as $\mathbf{v}$, described in Section \ref{sec:proxybce}.
This method is the state-of-the-art literature approach proposed by Haq \etal \cite{haq2021deep} in terms of model architecture and training, but does not include the nearest neighbour graph.

\subsubsection{Multi-label Proxy-NCA (\textit{ML-ProxyNCA}).} The standard Proxy-NCA \cite{movshovitz2017no, teh2020proxynca++} loss function can be naively extended to the multi-label case by optimising the following loss function:
\begin{equation} \label{proxynca}
    proxyNCA(\mathbf{v},\mathbf{y}) = -\log \left( \sum_{i=1}^C y_i \frac{ a_i}{\sum_{j=1}^C a_j}  \right),
\end{equation}
where $a_i = e^{-\|\mathbf{v}-\mathbf{p}_{i}\|^2 / 2\sigma^2}$ and $\mathbf{p}_{i}$ is the proxy for the i-th class. In the multi-label case, the outer sum over classes in (\ref{proxynca}) allows a training feature vector to pull towards the proxies belonging to all positive labels. In the case where each sample only has a single positive label, the equation in (\ref{proxynca}) becomes the standard probabilistic Proxy-NCA loss function \cite{teh2020proxynca++}.

{
\setlength{\tabcolsep}{6pt}
\begin{table}[!t]
\centering
\caption{CBIR performances of literature approaches, baselines and our approach, on the NIH dataset \cite{wang2017chestx}.}\label{NIHCompare}
\begin{tabular}{lc}
\toprule
Method & nDCG\\
\midrule
Wang \etal \cite{wang2012semi} & 0.15 \\
Gong \etal \cite{gong2012iterative} & 0.16\\
Erin Liong \etal \cite{erin2015deep} & 0.19\\
Liu \etal \cite{liu2016deep} & 0.17\\
Chen \etal \cite{chen2018order} & 0.24\\
Lan \etal \cite{lan2018simple} & 0.15\\
Haq \etal \cite{haq2021deep} & 0.31\\
DenseNet w/ BCE & 0.31 \\
ML-ProxyNCA & 0.32 \\
Ours & \textbf{0.38}  \\
\bottomrule
\end{tabular}
\end{table}
}

{
\setlength{\tabcolsep}{2pt}
\begin{table}[!t]
\centering
\caption{Classification and CBIR results on CheXpert \cite{irvin2019chexpert}.} \label{chexpertCompare}
\begin{tabular}{lcccc}
\toprule
 & Classification & \multicolumn{3}{c}{CBIR} \\
\cmidrule(lr){2-2} \cmidrule(lr){3-5}
Method & AUC & nDCG & ACG & Prec. \\
\midrule
DenseNet w/ BCE & 0.69 & 0.21 & 0.37 & 0.71\\
%Triplet & & & \\
ML-ProxyNCA & 0.64 & 0.20 & 0.36 & 0.72 \\
% \midrule
Ours (No Neg. Proxies) & 0.74 & 0.28 & 0.46 & 0.80 \\
Ours (With Neg. Proxies) & \textbf{0.77} & \textbf{0.30} & \textbf{0.48} & \textbf{0.82} \\
\bottomrule
\end{tabular}
\end{table}
}

\begin{figure*}[t!] 
    \centering
    \begin{subfigure}[t]{0.25\textwidth}
        \centering
        \includegraphics[width=\textwidth]{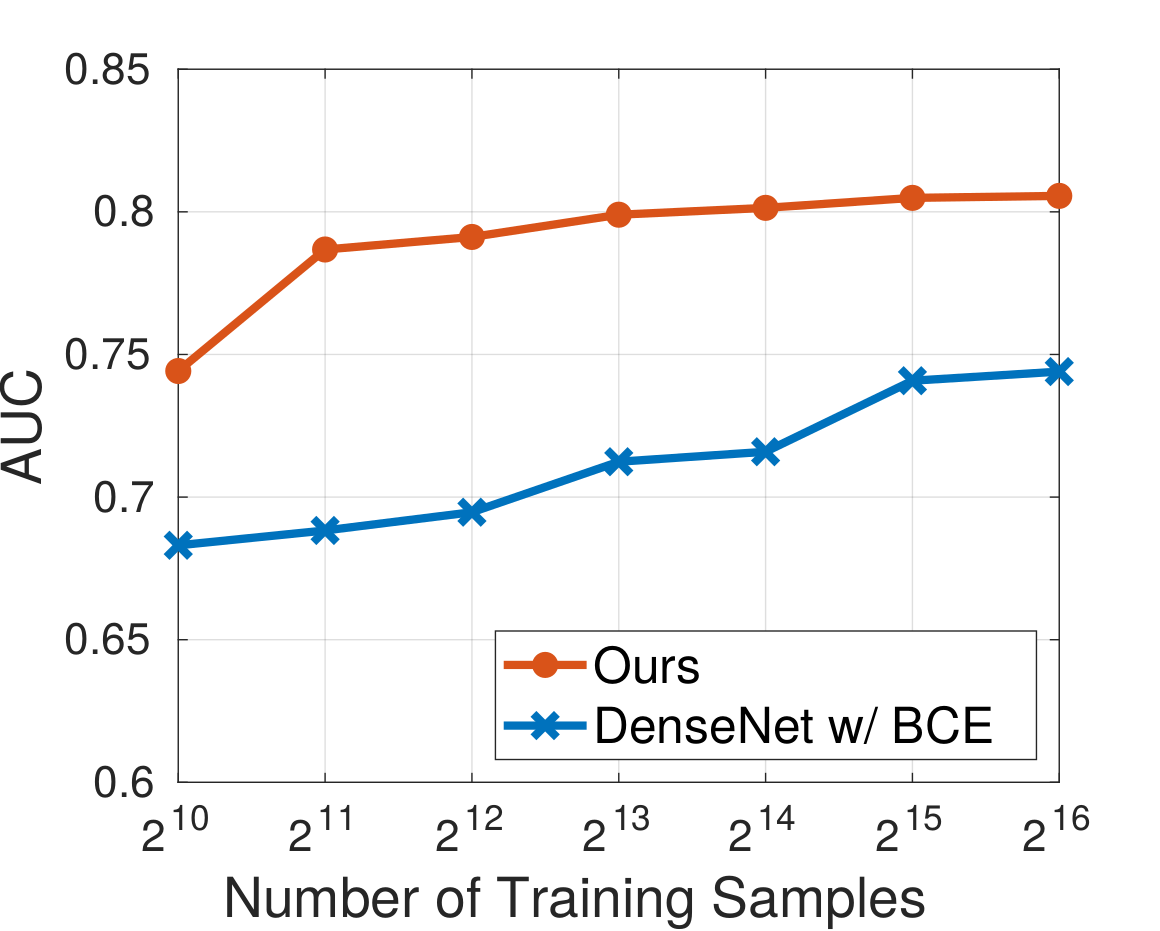}
        \caption{Classification AUC.}
    \end{subfigure}%
    \begin{subfigure}[t]{0.25\textwidth}
        \centering
        \includegraphics[width=\textwidth]{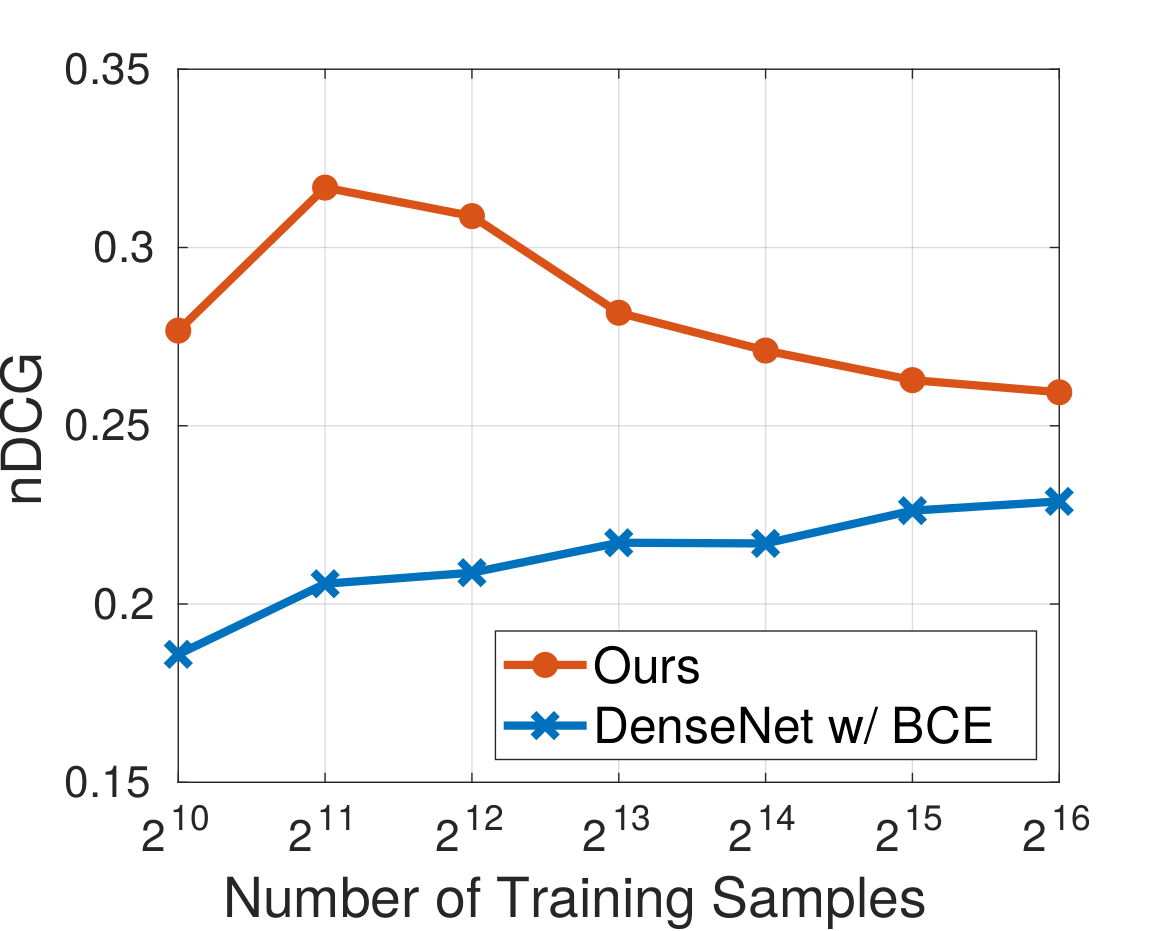}
        \caption{CBIR nDCG.}
    \end{subfigure}%
    \begin{subfigure}[t]{0.25\textwidth}
        \centering
        \includegraphics[width=\textwidth]{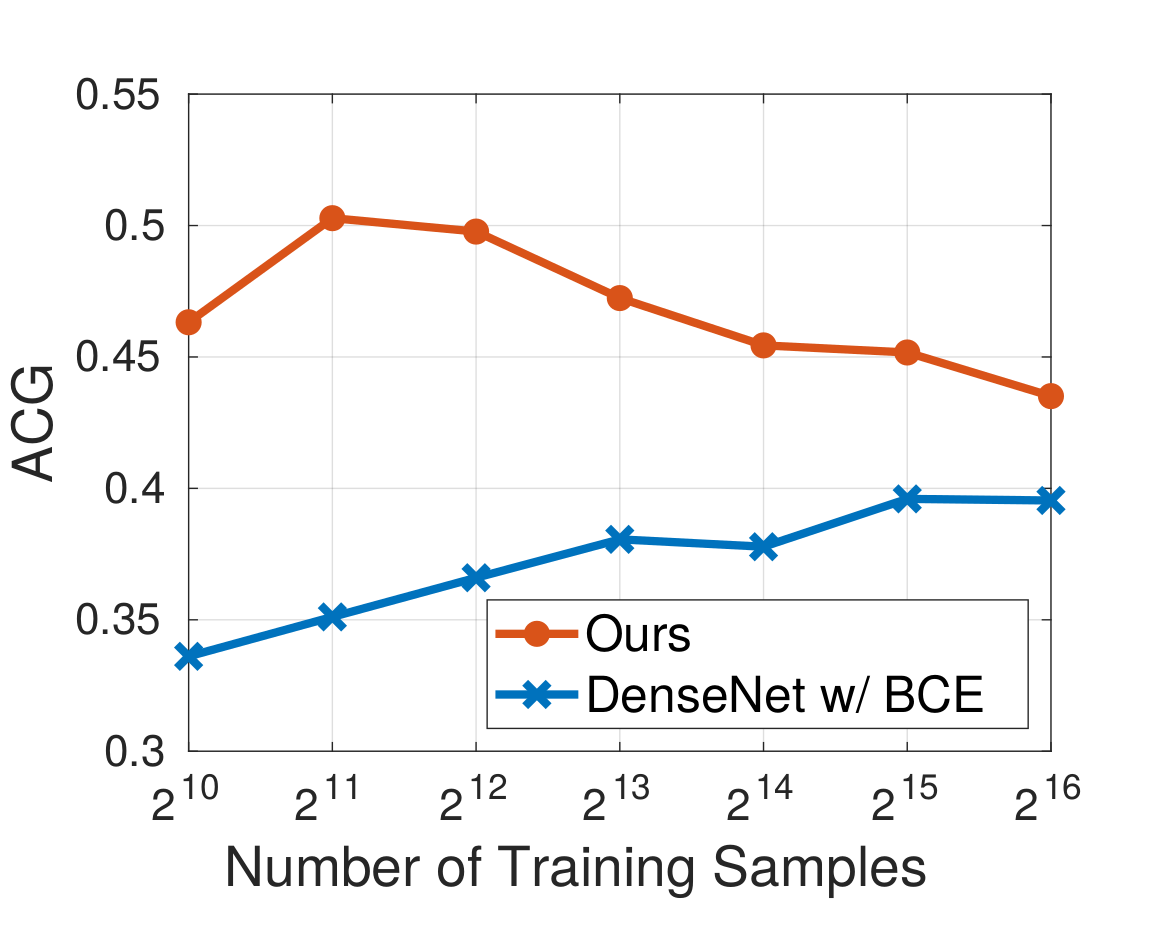}
        \caption{CBIR ACG.}
    \end{subfigure}%
    \begin{subfigure}[t]{0.25\textwidth}
        \centering
        \includegraphics[width=\textwidth]{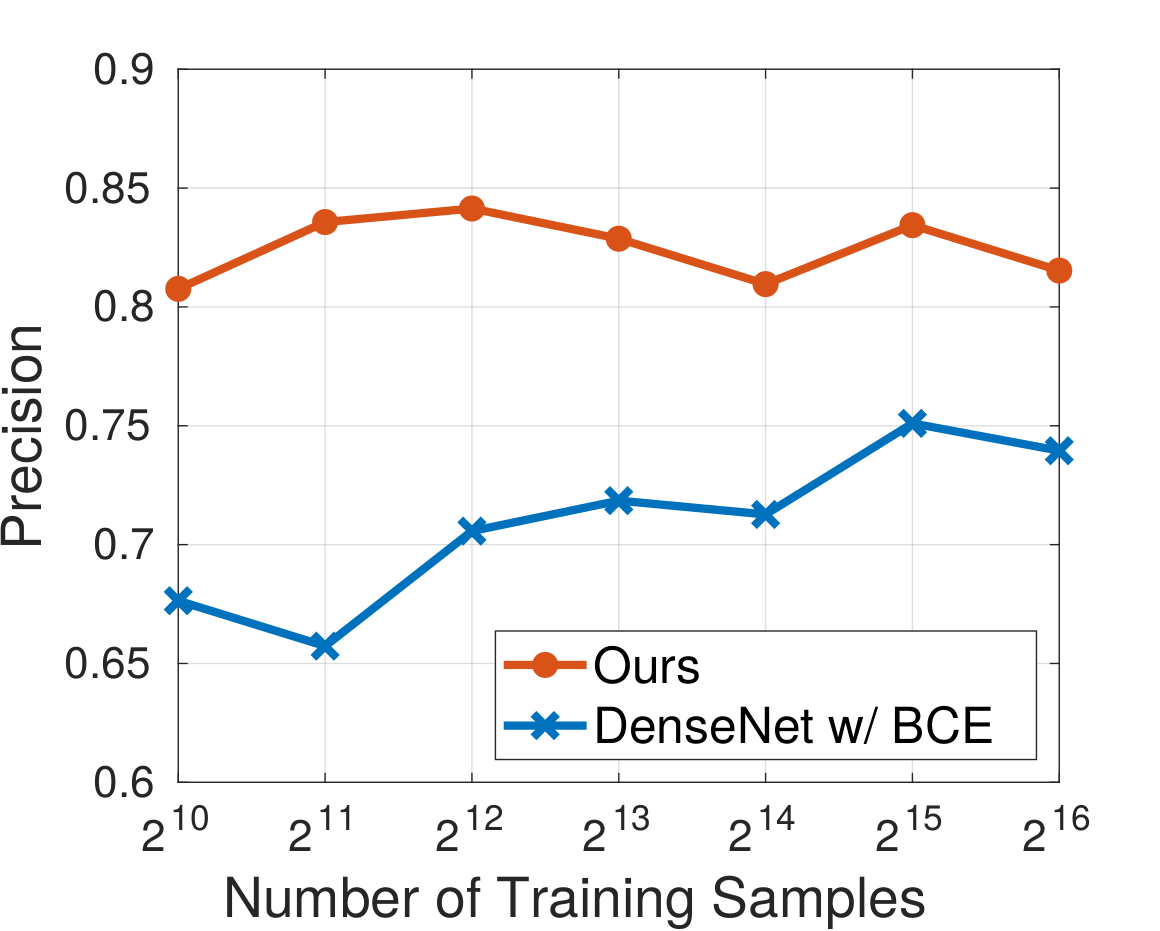}
        \caption{CBIR Precision.}
    \end{subfigure}
    \caption{Classification and image retrieval results across a range of dataset sizes. Negative proxies are used for our approach. Note the logarithmic scale of the horizontal axes. See Section \ref{sec:baselines} for details on baseline \textit{DenseNet w/ BCE}.}
    \label{fig:dataset_size}
\end{figure*}

\section{Experiments} \label{sec:exp}

\subsection{Implementation}
We use a DenseNet121 architecture \cite{huang2017densely} as our feature extractor network $f$, producing a feature vector $\mathbf{v}$ with 1024 dimensions. The Adam optimisation algorithm \cite{kingma2014adam} is used for model training, with coefficients $\beta_1 = 0.9$ and $\beta_2 = 0.999$. Our proposed method is trained for 50 epochs, with a learning rate of $10^{-4}$ and a batch size of 48. The loss function hyperparameter $\sigma$ is set to 0.7, and unless otherwise stated, two proxies are used per class. Hyperparameter values were tuned using a withheld validation set.

During training, images are first resized to 270x270 and then randomly cropped to 224x224. Finally, images are normalised to a range of -1 to 1. For evaluation images (retrieval database and queries), the random cropping is replaced by a centre crop. For evaluating content-based image retrieval, we compute the metrics using the $k$-nearest database samples for each query. 
For the CheXpert and NIH datasets, $k$ is set to 10 and 100, respectively. This evaluation protocol follows the set-ups used in Haq \etal \cite{haq2021deep} and Chen \etal \cite{chen2018order}.
In the interest of fairness, the same backbone network, data augmentation and data preprocessing is used for the baselines methods.

\subsection{Datasets} \label{sec:datasets}

CheXpert \cite{irvin2019chexpert} is comprised of frontal and lateral chest X-rays from 67,740 individual patients.
Following Haq \etal \cite{haq2021deep}, we use the nine most common diseases found in the dataset. The diseases and their shorthand abbreviations used in this paper are:
Enlarged Cardiomediastinum (EC),
Cardiomegaly (CM),
Lung Opacity (LO),
Edema (ED),
Consolidation (CS),
Pneumonia (PNA),
Atelectasis (AT),
Pneumothorax (PTX) and 
Pleural Effusion (PE).
Each sample is labelled as positive, negative or uncertain for each disease.
Uncertain labels are ignored during training, but treated as positive for CBIR purposes, following the set-up used by Haq \etal \cite{haq2021deep}.
The NIH Chest X-ray Dataset \cite{wang2017chestx} consists of frontal-view X-ray images from 30,805 patients. Again following Haq \etal \cite{haq2021deep}, we use the 13 most common disease labels in our experiments.

\begin{figure*}[t!] 
    \centering
    \includegraphics[width=0.9\textwidth]{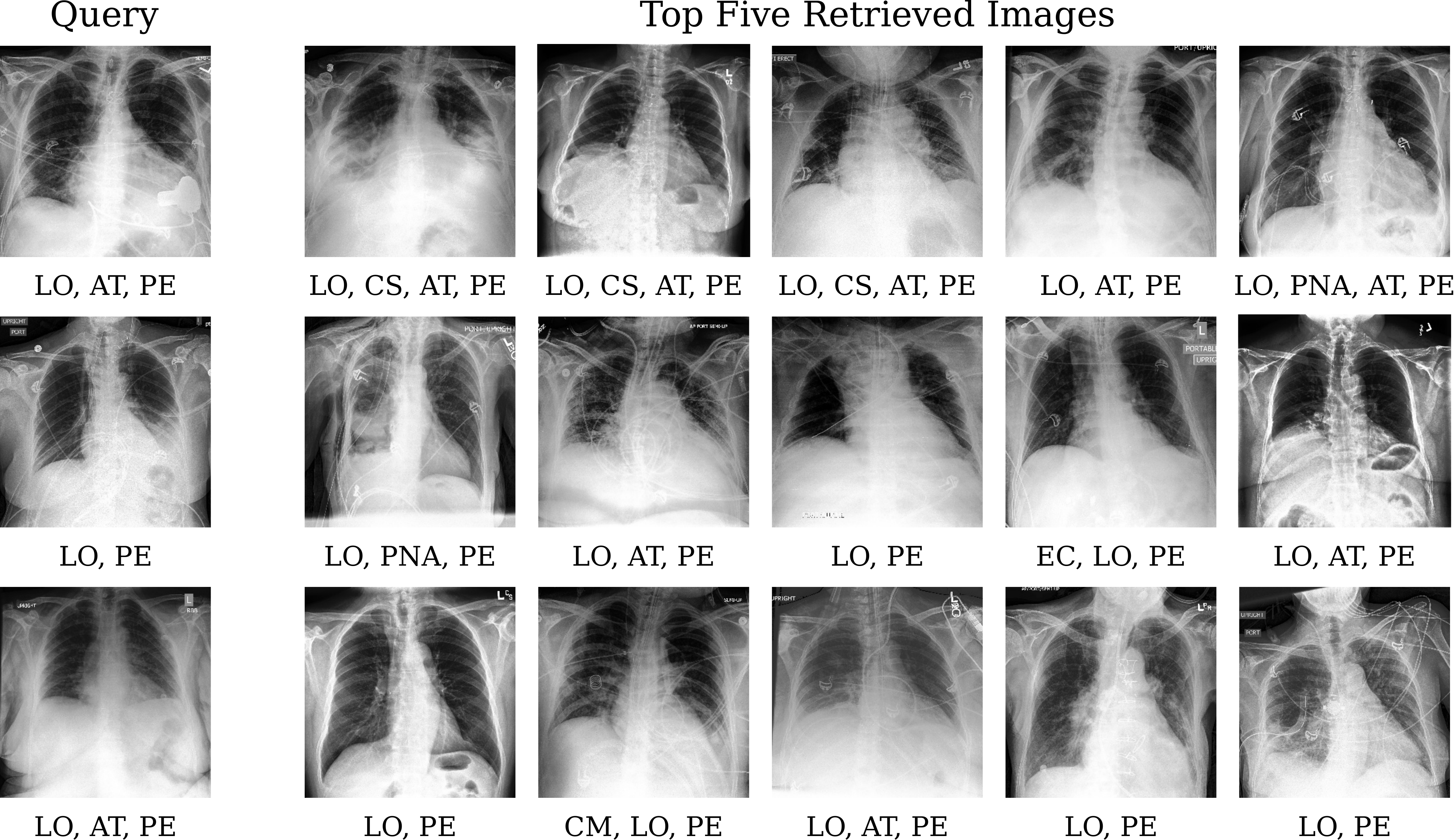}
    \caption{Qualitative CBIR results. Images are annotated with disease labels (see Section \ref{sec:datasets} for label definition).}
    \label{fig:cbir_result}
\end{figure*}

\subsection{Evaluation metrics}
For evaluating CBIR performance, we analyse the three retrieval metrics described below.
\subsubsection{Normalised Discounted Cumulative Gain (nDCG).}

Discounted Cumulative Gain is the sum of the graded relevance of all of the retrieved images based on their rank position:
\begin{equation}
    \mathrm{DCG} = \sum^k_{n=1}\frac{2^{r_n}-1}{\log(n+1)}
\end{equation}
where $k$ is the number of images that the system is retrieving and $r_n$ is the graded relevance value of the $n$-th retrieved image. Each graded relevance value is the number of common positive labels between the query and retrieved image. Each value is adjusted logarithmically proportional to the rank position $n$ of the query image. This means that a highly relevant query image will be penalised if it has a low rank.
The normalised DCG (nDCG) is defined as:
\begin{equation}
    \mathrm{nDCG} = \frac{\mathrm{DCG}}{\mathrm{iDCG}}
\end{equation}
where the ideal DCG (iDCG) is the maximum possible DCG that can be achieved based on the dataset.

\subsubsection{Average Cumulative Gain (ACG).} The ACG is defined as:
\begin{equation} 
    \mathrm{ACG} = \frac{\sum^k_{n=1}s_n}{k}
\end{equation}
where $s_n$ is the graded similarity value of the $n$-th retrieved image. Graded similarity is defined as the ratio of the number of common positive labels between the query image and the $n$-th retrieved image, and the total number of positive labels in the query.

\subsubsection{Precision (Prec.).} The CBIR precision is the ratio of the number of relevant images and the total number of retrieved images, $k$. Each retrieved image that has at least one common label with the query image is considered to be a relevant image. Precision is defined in (\ref{eq:precision}), where $\delta(.) \in \{0, 1\}$ is an indicator function.

\begin{equation} \label{eq:precision}
    \mathrm{precision} = \frac{\sum^k_{n=1}\delta(r_n>0)}{k}
\end{equation}

In order to evaluate pathology classification performance, we report the Area Under Receiver Operating Characteristic Curve (AUC). The receiver operating characteristic curve is a plot of the classifier's True Positive Rate (TPR) against the False Positive Rate (FPR), produced by sweeping the classifier's discrimination thresholds. The AUC measure will be between 0 and 1, where a higher value indicates a more robust classifier with a lower FPR across a range of TPRs.

\begin{figure*}[t!]
    \centering
    \begin{subfigure}[t]{0.25\textwidth}
        \centering
        \includegraphics[width=\textwidth]{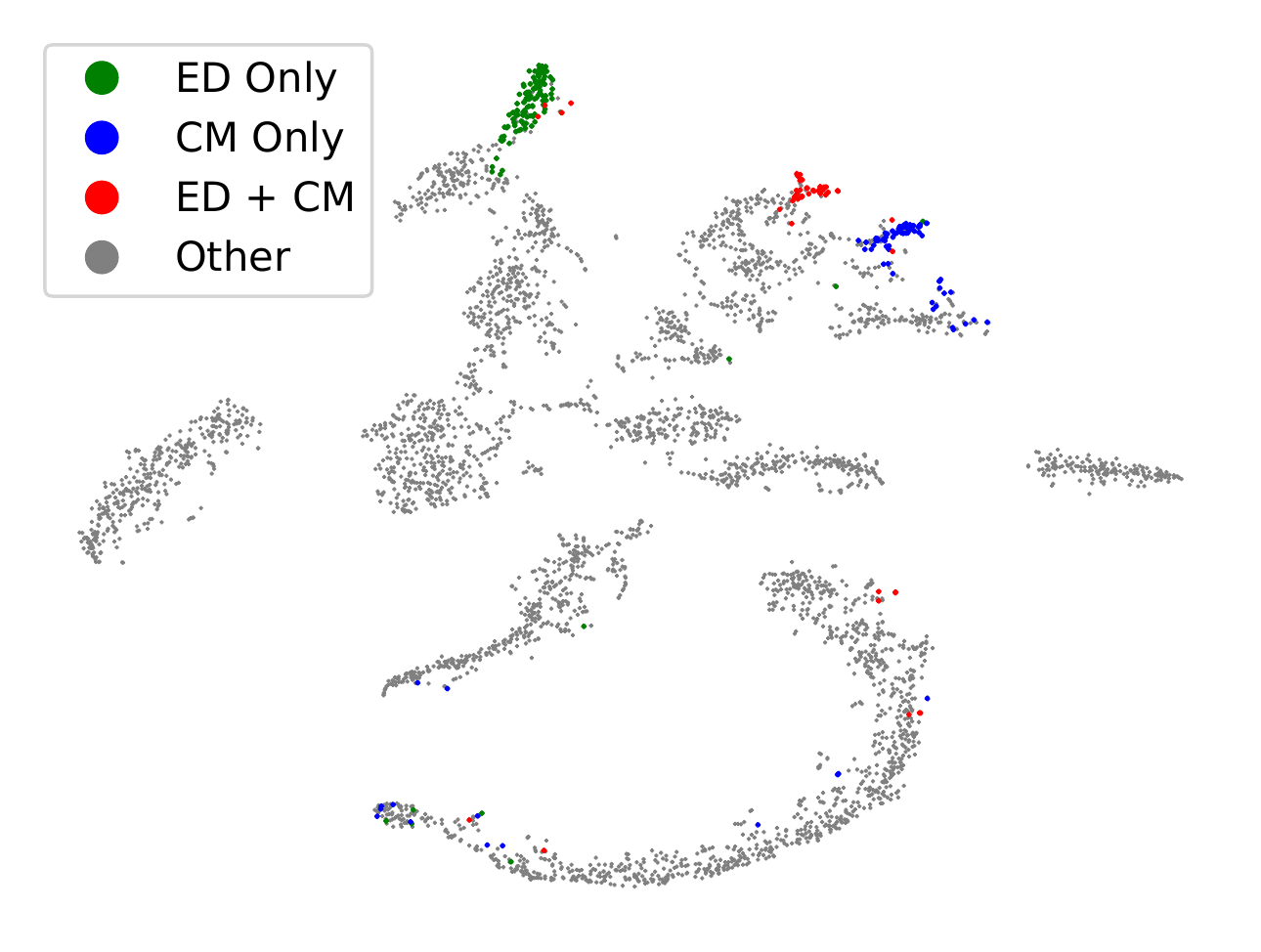}
        \caption{Edema/Cardiomegaly.}
    \end{subfigure}\hspace{1cm}%
    \begin{subfigure}[t]{0.25\textwidth}
        \centering
        \includegraphics[width=\textwidth]{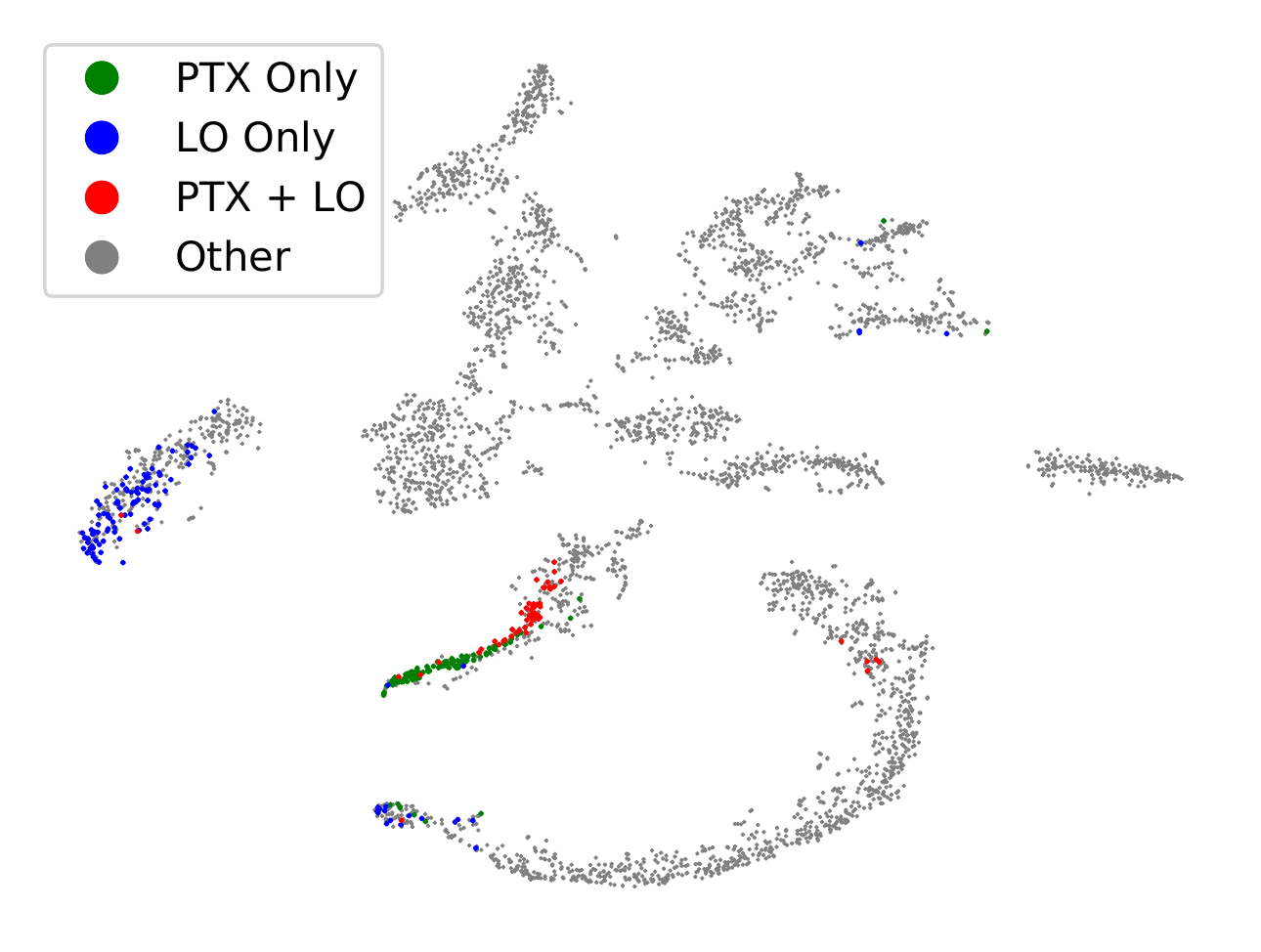}
        \caption{Pneumothorax/Lung Opacity.}
    \end{subfigure}\hspace{1cm}%
    \begin{subfigure}[t]{0.25\textwidth}
        \centering
        \includegraphics[width=\textwidth]{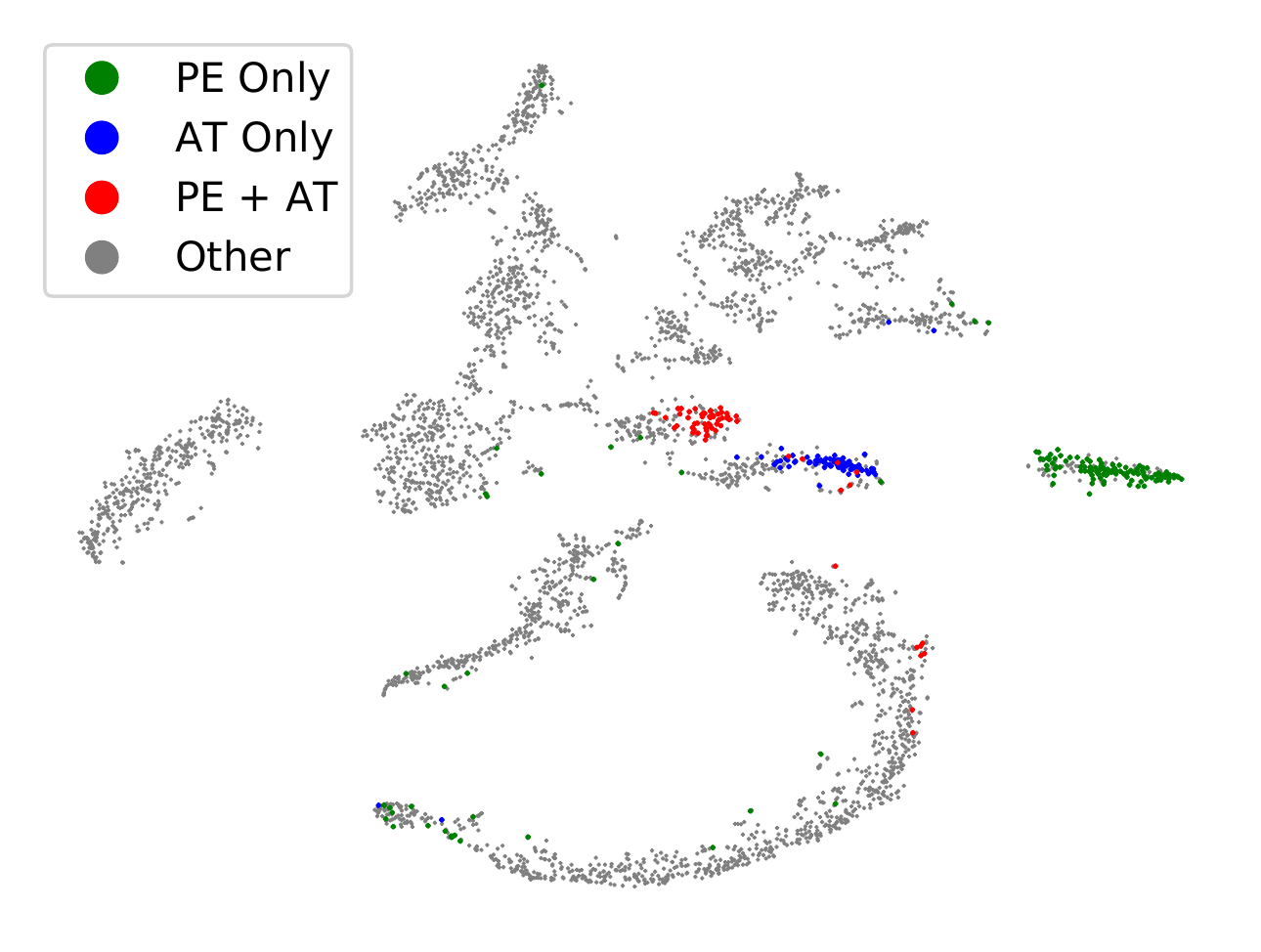}
        \caption{Pleural Effusion/Atelectasis.}
    \end{subfigure}
    \caption{Visualisation of the retrieval database feature vector space. Best viewed zoomed in on a monitor.}
    \label{fig:tsne}
\end{figure*}

\begin{figure*}[t!] 
    \centering
    \begin{subfigure}[t]{0.44\textwidth}
        \centering
        \includegraphics[width=\textwidth]{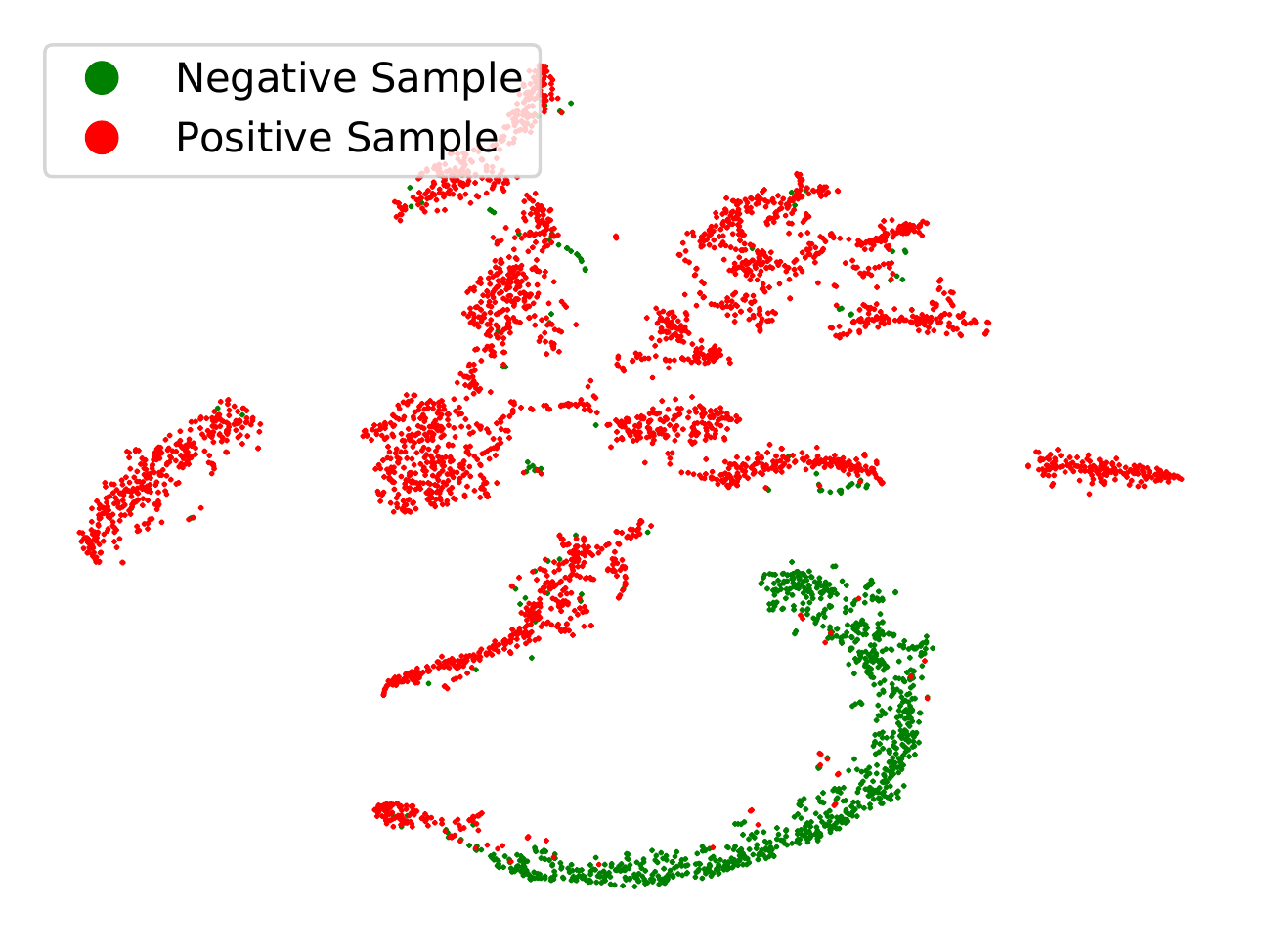}
        \caption{With negative proxies.}
    \end{subfigure}%
    \begin{subfigure}[t]{0.44\textwidth}
        \centering
        \includegraphics[width=\textwidth]{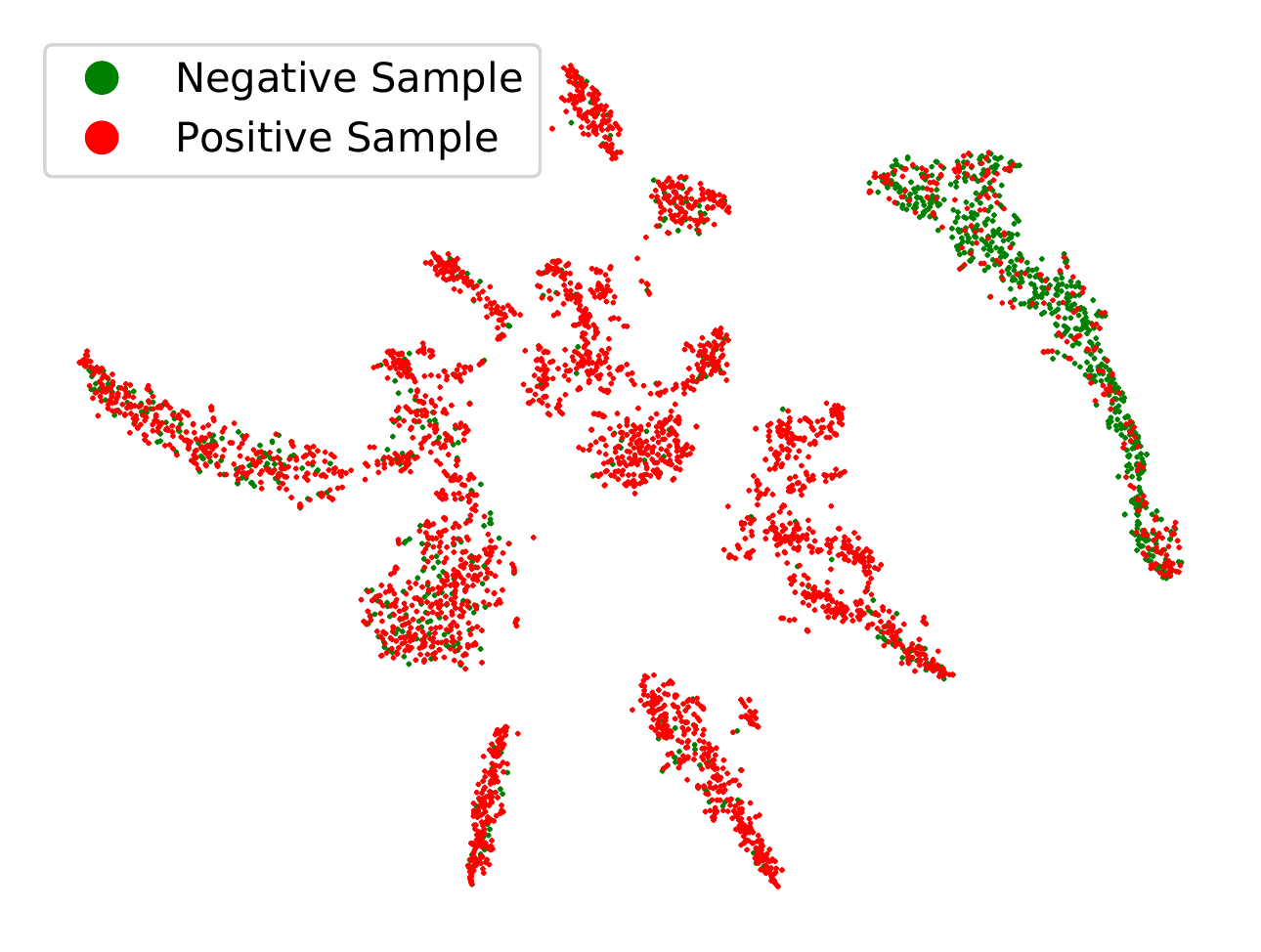}
        \caption{Without negative proxies.}
    \end{subfigure}
    \caption{Feature space visualisation showing the effect of including proxies for negative samples, i.e. samples with no positive labels. A \textit{positive sample} is a sample with any positive disease label. When negative proxies are used, negative samples are better clustered and are co-located with fewer positive samples. Best viewed zoomed in on a monitor.}
    \label{fig:negative_proxy_viz}
\end{figure*}

\begin{figure}[t!] 
    \centering
    \includegraphics[width=0.4\textwidth]{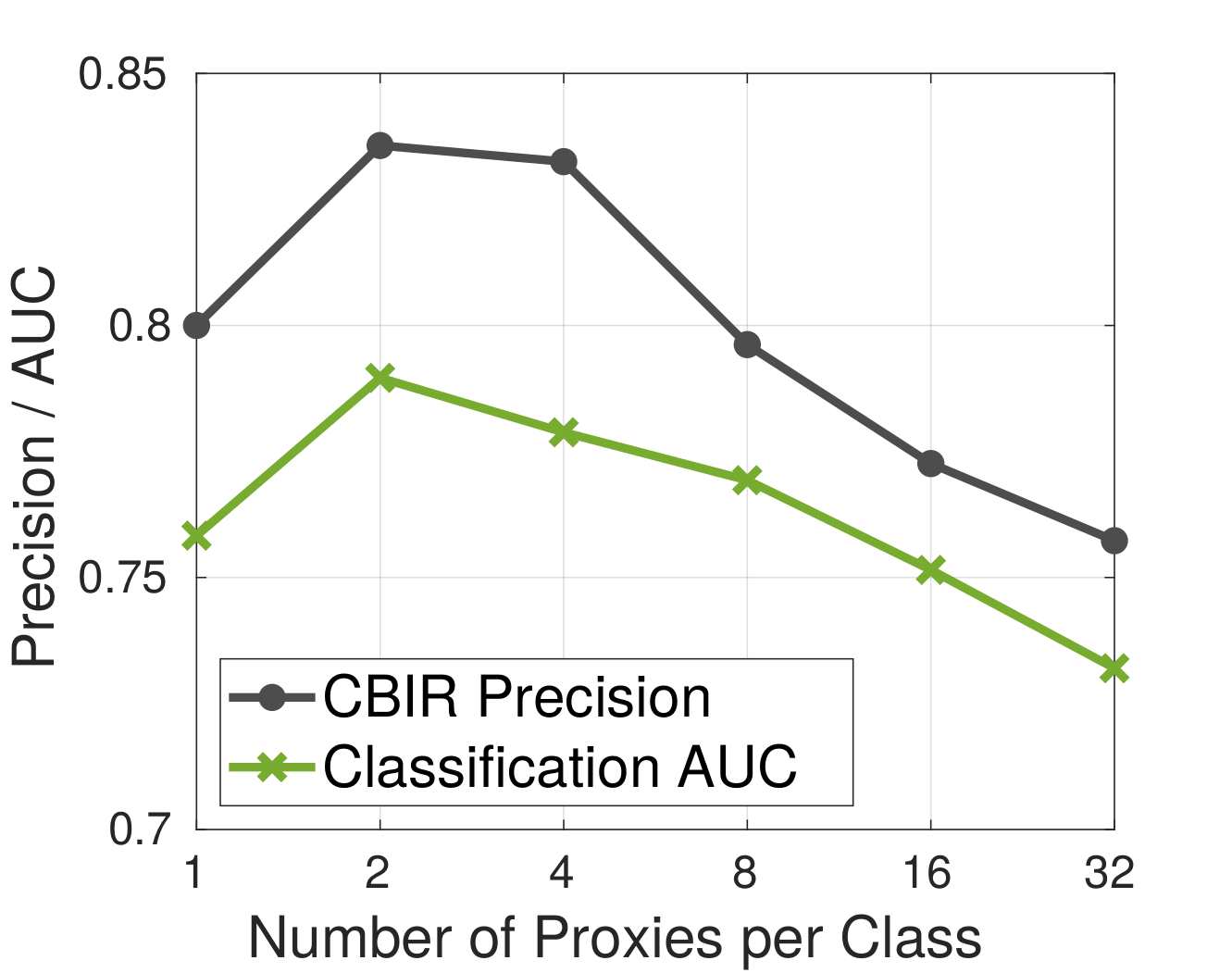}
    \caption{Effect of the number of proxies defined per class label on classification and image retrieval performance. Note the logarithmic scale of the horizontal axis.}
    \label{fig:num_proxies}
\end{figure}

\subsection{NIH literature comparison} \label{sec:nih_res}
We compare our approach to state-of-the-art image retrieval methods from literature on the NIH dataset. All of the compared methods were designed for multi-label data, while three of the methods are multi-label metric learning approaches \cite{chen2018order,liu2016deep,erin2015deep}. We follow the experimental set ups from Haq \etal \cite{haq2021deep} and Chen \etal \cite{chen2018order}. For training, 12,000 images are used, each containing at least one positive label from one of the 13 most common diseases in the dataset. As such, negative proxies are not used in this experiment. For testing, a further 1,000 samples are selected. As seen in the retrieval nDCG results in Table \ref{NIHCompare}, our method significantly outperforms the existing methods on the CBIR task.

\subsection{CheXpert evaluation} \label{sec:chexpert_res}
To further evaluate our method, we compare both classification and CBIR results on the CheXpert dataset to the baseline methods detailed in Section \ref{sec:baselines}. As training sample efficiency is important in the medical domain, due to the difficulty of obtaining high-quality annotations, we evaluate performance across a range of training set sizes, with a particular interest in the smaller sizes. In these experiments, both positive samples (with at least one positive label from one of the nine most common diseases) and negative samples (no positive labels) are used. As such, negative proxies are leveraged in these experiments. 

Table \ref{chexpertCompare} shows all classification and CBIR metrics using a training set size of 4096 samples. Our approach outperforms the baselines methods across all classification and CBIR metrics.
The comparatively poor performance of the naive multi-label extension of Proxy-NCA (ML-ProxyNCA) in Tables \ref{NIHCompare} and \ref{chexpertCompare} demonstrates that multi-label proxy metric learning is non-trivial to achieve. The naive extension performs comparatively poorly at both pathology classification and CBIR, while our novel formulation of proxy metric learning for multi-label data is highly effective for both tasks. 

Fig. \ref{fig:dataset_size} compares performance of our method to the classifier baseline (DenseNet w/ BCE) across a large range of dataset sizes (from 1024 - 65536 samples). These results show a consistent and significant performance advantage to our method in both pathology classification and CBIR. The advantage holds across both small and large training set sizes, and is largest when training data is limited. This shows the suitability of our method to data constrained medical problems.

\subsection{Qualitative evaluation}
Example image retrieval results for sample query X-rays are shown in Fig. \ref{fig:key_result} and \ref{fig:cbir_result}. Query images are from a test set that is withheld during training. Disease annotations are indicated by the shorthand names defined in Section \ref{sec:datasets}. In general, there is a strong similarity between the disease labels of the query samples and retrieved samples. Fig. \ref{fig:tsne} uses t-SNE visualisations \cite{van2008visualizing} of the retrieval database feature vector space to show some of the relationships between diseases learned by the model. Each visualisation selects two disease labels and indicates by colour the samples that have only one of those labels positive or both labels positive. Samples are co-located in the feature space based on their combination of positive disease labels.

\subsection{Proxies per class} \label{sec:num_proxies}
Fig. \ref{fig:num_proxies} shows the effect that varying the number of proxies defined for each class label has on the classification and CBIR performance. There is a benefit to having multiple proxies per class, with two providing the best results. Interestingly, as the proxies per class passes eight, the performance begins to drop below the level of a single proxy per class.
This is likely due to over-parameterisation of the feature space resulting in overfitting and proxies that do not generalise as well to unseen samples. For example, the extreme case of this would be having one proxy for each training sample, where without additional regularisation, the model wouldn't necessarily need to generalise.

\subsection{Negative proxies} \label{sec:neg_proxy_res}
We analyse the effect of using negative proxies both quantitatively and qualitatively. Table \ref{chexpertCompare} shows that excluding negative proxies during training results in a performance drop for both classification and CBIR.
The t-SNE visualisation \cite{van2008visualizing} of the feature vector space in Fig. \ref{fig:negative_proxy_viz} shows that without negative proxies, more positive samples are peppered throughout the primary negative sample cluster, compared to when negative proxies are used.
Negative proxies are important for accurately encoding the semantic content of samples with no disease findings, and allow for better discrimination between negative and positive samples.

\section{Conclusion and future work}
Computer-aided diagnosis systems can help to reduce the workload of healthcare professionals, potentially resulting in improved patient outcomes \cite{haq2021deep}. In this paper, we presented a novel model that can be jointly used for pathology classification and content-based image retrieval. Our multimorbidty metric learning approach uses the power of proxies to efficiently learn a feature vector space that encodes the relationships between disease labels. We showed the efficacy of our approach on two chest X-ray datasets, demonstrating a performance advantage over the baseline and state-of-the-art methods in classification and retrieval.

Defining multiple proxies for each disease resulted in improved classification and CBIR performance. This leads to a research question: can defining a variable number of proxies across diseases help to alleviate the affects of unbalanced data? Such unbalanced data is common in the medical domain, where particular disease annotations may be scarce due to the rarity of the disease. Setting per disease proxy numbers with consideration to the disease distribution may help to improve model training on unbalanced data. We leave this as a promising future research direction.

\bibliographystyle{unsrt}

\end{document}